%% file: iclr2021_RL_flexibility.tex
\documentclass{article} 
\usepackage{iclr2021_conference,times}

\input{math_commands.tex}
\usepackage{hyperref}       
\usepackage{url}            
\usepackage{booktabs}       

\usepackage{bm}
\usepackage{multirow}
\usepackage{amsmath}
\usepackage{amssymb}
\usepackage{amsthm}
\usepackage{color}
\usepackage{times}
\usepackage{graphicx} 
\usepackage{subcaption}
\usepackage[rightcaption]{sidecap}
\usepackage{threeparttable}
\usepackage{algorithm}
\usepackage{algorithmic}
\usepackage{wrapfig}

\ifx\remark\undefined
\newtheorem{remark}{Remark}
\fi

\title{Reinforcement Learning for Flexibility Design Problems}

\author{Yehua Wei \thanks{ Equal contribution. Code for RL implementation: https://github.com/zhlzhl/RL\_flex\_design} \\
Duke University \\
\And
Lei Zhang \footnotemark[1] \\
Fidelity Investments\\
\And
Ruiyi Zhang \\
Duke University \\
\And
Shijing Si \\
Duke University \\
\And
Hao Zhang \\
Cornell University \\
\And
Lawrence Carin \\
Duke University \\

}

\iclrfinalcopy 
\begin{document}

\maketitle

\begin{abstract}
Flexibility design problems are a class of problems that appear in strategic decision-making across industries, where the objective is to design a ($e.g.$, manufacturing) network that affords flexibility and adaptivity. The underlying combinatorial nature and stochastic objectives make flexibility design problems challenging for standard optimization methods. In this paper, we develop a reinforcement learning (RL) framework for flexibility design problems. Specifically, we carefully design mechanisms with noisy exploration and variance reduction to ensure empirical success and show the unique advantage of RL in terms of fast-adaptation. Empirical results show that the RL-based method consistently finds better solutions compared to classical heuristics.
\end{abstract}

\section{Introduction}


Designing flexibility networks (networks that afford the users flexibility in their use) is an active and important problem in Operations Research and Operations Management \citep{chou2008process, wang2019review}. The concept of designing flexibility networks originated in automotive manufacturing, where firms considered the strategic decision of adding flexibilities to enable manufacturing plants to quickly shift production portfolios with little penalty in terms of time and cost \citep{JG95}. \citet{JG95} observed that with a well designed flexible manufacturing network, firms can better handle uncertainties and significantly reduce idling resources and unsatisfied demands.

\begin{figure}[t!]
\centering
 \begin{subfigure}[t]{0.4\textwidth}
 \centering
 \includegraphics[scale=0.2]{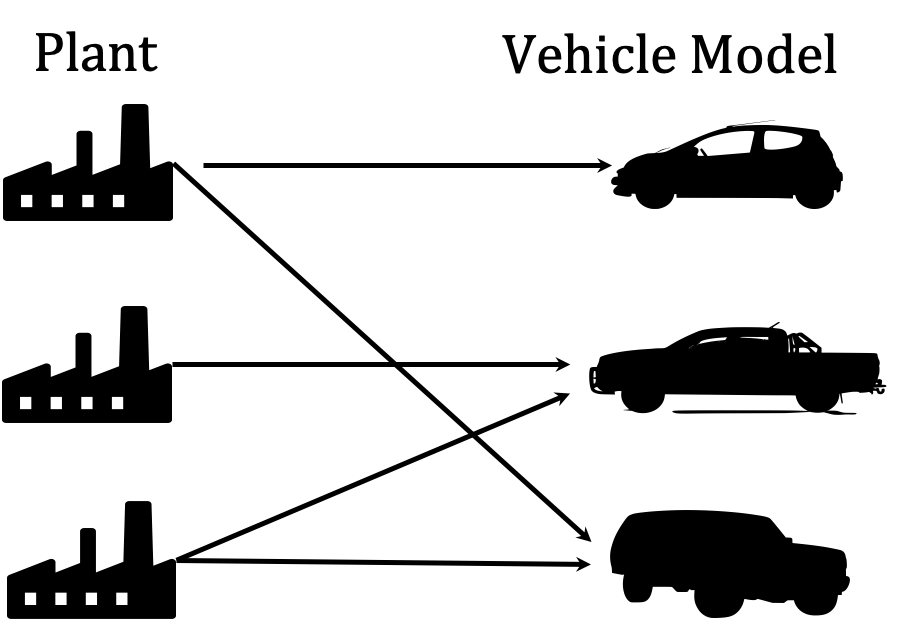}
   \caption {network for an auto-manufacturer}\label{fig:flex_auto}
 \end{subfigure}
 ~
 \begin{subfigure}[t]{0.5\textwidth}
  \centering
  \includegraphics[scale=0.2]{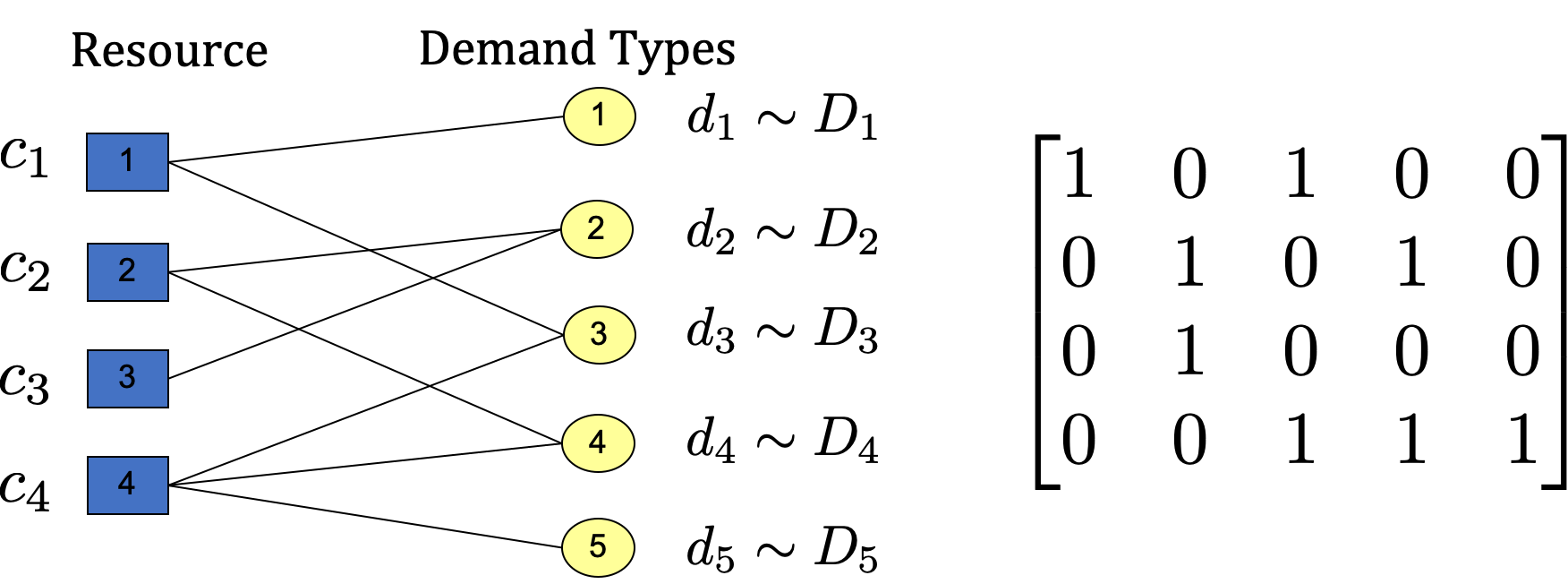}
     \caption {generic network and its matrix representation}\label{fig:flex_general} 
 \end{subfigure}
\caption {\textit{Examples of Flexibility Networks}  \label{fig:flex_examples}}
\vspace{-6mm}
\end{figure}

With the increasingly globalized social-media-driven economy, firms are facing an unprecedented level of volatility and their ability to adapt to changing demands has become more critical than ever. As a result, the problem of designing flexibility networks has become a key strategic decision across many different manufacturing and service industries, such as automotive, textile, electronics, semiconductor, call centers, transportation, health care, and online retailing \citep{chou2008process, simchi2010operations, wang2015process, asadpour2016online, devalve2018understanding, ledvina2020new}. The universal theme in all of the application domains is that a firm has multiple resources, is required to serve multiple types of uncertain demands, and needs to design a flexibility network that specifies the set of resources that can serve each of its demand types. For example, an auto-manufacturer owns multiple production plants (or production lines), produces a portfolio of vehicle models, and faces stochastic demand for each of its vehicle models. The firm's manufacturing flexibility, coined as process flexibility in \citet{JG95}, is the capability of quickly switching the production at each of its plants between multiple types of vehicle models. As a result, the firm needs to determine a flexible network, where an arc $(i,j)$ in the network represents that plant $i$ is capable of producing a vehicle model $j$ (see Figure \ref{fig:flex_auto} for an example). In another example, a logistics firm owns multiple distribution centers and serves multiple regions with stochastic demand. To improve service quality and better utilize its distribution centers, the firm desires an effective flexible network with arcs connecting the distribution centers and regions, where each arc represents an investment by the firm to enable delivery from a distribution center to a region.



In general, the flexibility design problem (FDP) is difficult to solve. It is a combinatorial optimization problem with stochastic demand, that is already NP-hard even when the demand is deterministic. The stochastic nature of the FDP eliminates the possibility of finding the optimal solution for medium-sized instances with $m \cdot n\geq 50$, where $m$ and $n$ denote the number of resources and demand types, even when stochastic programming techniques such as Benders Decomposition are applied \citep{feng2017process}. Therefore, researchers have mostly focused on finding general guidelines and heuristics \citep{JG95, Chou2010, SimchiLeviWei, feng2017process}. While the heuristics in literature are generally effective, in many applications, even a one percent gain in a FDP solution is highly significant \citep{JG95, Chou2010, feng2017process} and this motivates us to investigate reinforcement learning (RL) as a new approach for solving FDPs. In addition, FDPs fall into the class of two-stage mixed-integer stochastic programs, which often arise in strategic-level business planning under uncertainty (see, $e.g.$, \citet{santoso2005stochastic}). Therefore, RL's success in FDPs can lead to future interest in applying RL to more strategic planning problems in business. We believe that RL can be effectively applied in FDPs and more strategic decision problems for two reasons. First, RL is designed to solve problems with stochastic rewards, and the stochasticity of strategic planning problems may even help RL to explore more solutions~\citep{sutton2000policy}. Secondly, the heuristics are mostly relied on, and therefore, constrained by human intuitions, whereas RL algorithms with deep neural networks may uncover blind spots and discover new designs, or better, new intuition altogether similar to the successful applications of RL in Atari games and AlphaGo Zero \citep{silver2017mastering}.

In this paper, we develop an RL framework to solve a given FDP. We first formulate an FDP as a Markov Decision Process (MDP), and then optimize it via policy gradient. In this framework, we design a specific MDP to take advantage of the structure possessed by FDPs. Further, a problem-dependent method for estimating a stochastic objective, $i.e.$, terminal rewards, is proposed. 
Empirical results show that our proposed method consistently outperforms classical heuristics from the literature. Furthermore, our framework incorporates fast-adaptation for similar problems under different settings, and hence, avoids repeated computations as in classical heuristics.




\section{Flexibility Design Problem} 

We present the formal formulation for the FDP. In an FDP, there are $m$ resources (referred to as supply nodes) and $n$ demand types (referred to as demand nodes). The set of feasible solutions $\mathcal{F}$ is all networks with no more than $\K$ arcs. Formally, each network $\flex$ is represented by an $m\times n$ matrix (see Figure \ref{fig:flex_general}), implying that
$\mathcal{F} = \{\flex \,|\, \flex \in \{0,1\}^{m n},\, \sum_{i\in[m],j\in[n]}F_{ij} \leq \K\},$
where $[m]$ and $[n]$ denote the set of integers from 1 to $m$ and 1 to $n$, respectively. 


The objective of FDP is to find the flexibility network $\flex \in \mathcal{F}$ that maximizes the expected profit under stochastic demand. Denoting the random demand vector as $\bd \in \R^n$ with distribution $\mathbf{D}$, the FDP can be defined as
\begin{equation}\label{eq:fdp}
    \max_{\flex \in \mathcal{F}} \Ex_{\bd \sim \mathbf{D}}[P(\bd, \flex)] - \sum_{i,j} I_{ij}F_{ij},
\end{equation}
where $I_{ij} \geq 0$ is the cost of including arc $(i,j)$ to the flexibility network, while $P(\bd, \flex)$ is the profit achieved by network $\flex$ after the demand is revealed to be $\bd$. It is important to note here that the decision on $\flex$ is made \emph{prior} to the observation of the demand, and thus, the decision is to create $\flex$ such that the expected profit (over $\bd \sim \mathbf{D}$) is optimized. We next define $P(\bd, \flex)$.

Given parameter $\bp \in \R^{mn}$ where $p_{ij}$ represents the unit profit of using one unit of resource $i$ to satisfy one demand type $j$ and $\bc \in \R^n$ where $c_i$ represents the capacity of resource $i$, the function $P(\bd, \flex)$ is defined as the objective value of the following linear programming (LP) problem:
\begin{equation*} \max_{\mathbf{f} \in \R^{m n}} \sum_{i,j}p_{ij}f_{ij},
 \mbox{s.t.} \, \sum_{i \in [m]}f_{ij} \leq d_j, \forall j \in [n], \sum_{j \in [n]}f_{ij} \leq c_i, \forall i \in [m],  0 \leq f_{ij} \leq M_{ij}F_{ij} , \forall i,j, 
\end{equation*}
where $M_{ij}$ is a large constant ensuring that the constraint $f_{ij} \leq M_{ij}F_{ij}$ is nonbinding when $F_{ij} = 1$. It is easy to check that it is sufficient to set $M_{ij} = \min\{c_i, d_j\}$ (see \citet{feng2017process} for a discussion). Intuitively, the LP defining $P(\bd, \flex)$ solves the optimal resource allocation subject to capacity for resource $i$ being $c_i$ and demand for type $j$ being $d_j$, while only using arcs in $\flex$. In general, the LP can be solved efficiently (in a fraction of a millisecond) by LP solvers such as Gurobi \citep{gurobi}, and we exploit this in our RL framework.


While $P(\bd, \flex)$ is easy to compute, solving FDP defined in \eqref{eq:fdp} is difficult. Even when $\bd$ is deterministic, FDP reduces to an NP-hard problem known as the fixed charge transportation problem (see, $e.g.$, \citet{agarwal2012fixed}). 
For the FDP with stochastic $\bd$, FDP is significantly more difficult than the typical NP-hard combinatorial optimization problems, such as travelling salesman or vertex cover problems, because there is no known method for even computing the objective function exactly except for some special cases \citep{Chou2010}. As a result, researchers mostly resort to estimating the objective by repeatedly drawing samples of $\bd$. While the estimation is accurate when the number of samples of $\bd$ is large, it reformulates FDP as a mixed-integer linear program (MILP) with a large number of variables and constraints, rendering most of the MILP algorithms impractical in this context.


\section{Related Work}

The research on designing flexibility networks can be divided into three streams that often overlap with each other: ($i$) extending its model and applications, ($ii$) conducting theoretical analysis, and ($iii$) proposing methods to solve FDPs (see \citet{chou2008process, wang2019review} for more details). The third stream of the literature is most related to our work. Due to the inherent complexity of the FDP, researchers so far have focused on heuristics instead of exact algorithms. Even for the classical flexibility design problem (assuming $p_{ij}=1$ and $I_{ij}=0$ for all $i,j$), the optimal solution of FDP is not known except for very restrictive cases \citep{desir2016sparse}. As a result, various heuristics have been proposed for the classical FDP \citep{Chou2010, Chou2011, simchi2015worst}. However, while the classical FDP is important, it has been observed that arc costs ($I_{ij}$) and different unit profits ($p_{ij}$) can be key determinants to the value of a flexibility network \citep{van1998investment, wang2019review}.
For general FDPs, the most intuitive heuristic 
is perhaps the greedy heuristic, which at each iteration, adds an arc with the highest performance improvement. The performance guarantee on such heuristics without fixed arc costs is derived in \citet{devalve2020matching}.
Alternatively, the heuristic proposed by \citet{feng2017process} solves a stochastic programming (SP) that relaxes the integer constraint of FDP at each iteration, and uses the fractional solution to make a decision on which arc to add. In addition, \citet{chan2019sparse} proposed a two-step procedure which first trains a neural net for predicting the expected profit of flexibility designs, then applies greedy-type heuristics to optimize flexibility designs based on the predictor.



Recently, there have been various activities in applying the recent advances in ML tools to solve combinatorial optimization problems, such as the traveling salesman problem (TSP) \citep{bello2016neural,vinyals2015pointer}, vehicle routing problem (VRP) \citep{kool2018attention, nazari2018reinforcement}, min vertex cover problem, and max cut problem \citep{abe2019solving}. Interested readers are referred to \citet{bengio2018machine} for a framework on how combinatorial problems can be solved using an ML approach and for a survey of various combinatorial optimization problems that have been addressed using ML. To the best of our knowledge, FDPs do not possess the property that solutions are naturally represented as a sequence of nodes, nor the property that the objective can be decomposed into linear objectives of a feasible integer solution. Hence, it is not straightforward to apply the existing RL works in combinatorial problems to FDPs.
In addition, Neural Architecture Search (NAS)~\citep{baker2016designing,zoph2016neural} shares some similarity to FDPs, where it aims at automatically searching for a good neural architecture via reinforcement learning to yield good performance on datasets such as CIFAR-10 and CIFAR-100. \citet{liu2017hierarchical} imposed a specific structure as the prior knowledge to reduce the search space and \citet{baker2017accelerating} proposed performance prediction for each individual architecture.



\section{Proposed Method}\label{sec:framework}
We present a reinforcement learning framework for flexibility designs. It is typically intractable to perform supervised training since FDPs are NP-hard and the objective is stochastic. However, reinforcement learning (RL) can provide a scalable and fast-adaptable solution based on the partial observation from the performance of a specific FDP solution. We first formulate a given FDP as a finite-horizon RL problem, then introduce policy optimization, evaluation and fast adaptation.

\subsection{FDPs as Reinforcement Learning}\label{sec:MDP}
The FDP can be considered as an RL problem, with discrete actions, deterministic state transitions, finite horizon and stochastic terminal rewards. 
For a given FDP defined in \eqref{eq:fdp},  an agent will sequentially select at most one arc at a step based on the current state for $\K$ discrete steps.
Accordingly, we can model the FDP as an MDP $\mathcal{M} = \langle\mathcal{S}, \mathcal{A}, P, R \rangle$, where ${P}:~\mathcal{S}\times\mathcal{A}\times\mathcal{S}\mapsto \mathbb{R}$ is the environment dynamic of the given FDP and $R: \mathcal{S}\times\mathcal{A}\mapsto \mathbb{R}$ is the stochastic reward function used to evaluate different arcs.
%
%
%
%
We denote $\mathcal{S}$ as the state space, which in our setting is the set of all flexibility networks connecting supply and demand nodes, $i.e.$, $\mathcal{S} = \{\flex \,|\, \flex \in \{0,1\}^{mn} \}$. 
We denote $\mathcal{A}$ as the action space, and  $\mathcal{A}=\{(i,j) \,|\, i\in[m], j\in[n] \}$.  Given state $\sv_t = \flex$, an action $\av = (i,j)\in\mathcal{A}$, the deterministic transition is given by $\sv_{t+1} = \flex \cup (i,j)$, where $\flex \cup (i,j)$ represents the network with all arcs of $\flex$ and arc $(i,j)$.

At each time step $t$, the agent observes the current flexibility network $\sv_t \in \mathcal{S}$, chooses an action $\av_t$, then receives an immediate reward. The initial state, $\sv_0$, is defined as the matrix with all zeros, $i.e.$, the network with no arcs. The objective of RL is to maximize the expected return as defined in \eqref{eq:rlobj}:
\begin{equation}
\begin{aligned}\label{eq:rlobj}
J(\pi) = \mathbb{E}_{\tau\sim p_\pi (\tau)}\left[\sum_{t=0}^K \gamma^t r(\sv_t, \av_t)\right]\,,\text{where}~r(\sv_t, \av_t) = \left\{
   \begin{array}{@{}ll@{}}
     I(\sv_t, \av_t), &\text{if}\ t<\K, \\
     \Ex_{\mathbf{d}\sim \mathbf{D}}[P(\bd, \sv_t)], &\text{if}\ t=\K,
   \end{array}\right.,
\end{aligned}
\end{equation}
where the trajectory $\tau=(\sv_0, \av_0,\ldots,\av_{K-1}, \sv_K)$ 
is a sequence of states and actions, the trajectory distribution $p_\pi(\tau):=\prod_{t=0}^K \pi(\av_t|\sv_t)T(\sv_{t+1}|\sv_t, \av_t)$, $I(\sv_t, \av_t) = -I_{ij}$ if $\av_t$ is adding arc $(i,j)$ to $\sv_t$ and $(i,j)\notin \sv_t$, and $I(\sv_t, \av_t) =0$ otherwise. The goal of an agent is to learn an optimal policy that maximizes $J(\pi)$. 


\vspace{-3mm}
\paragraph{Optimal Solution}
Considering the designed rewards in \eqref{eq:rlobj}, we have the Bellman equation as follows:
\begin{equation}\label{eq:bellman_a}
   V_{\pi}(\sv_t) = 
   \left\{
   \begin{array}{@{}ll@{}}
     \mathbb{E}_{\pi} \big[ I(\sv_t, \av_t)+ \gamma V_{\pi} (\sv_{t+1}) \big], & \text{if}\ t<\K, \\
     \Ex_{\mathbf{d}\sim \mathbf{D}}[P(\bd, \sv_t)], & \text{if}\  t = \K.
   \end{array}\right.
 \end{equation}
 
Note that \eqref{eq:bellman_a} has an optimal deterministic policy, and the optimal policy returns an optimal solution for the FDP defined in \eqref{eq:fdp}.

We next discuss sample-based estimation for $\Ex_{\mathbf{d}\sim \mathbf{D}}[P(\bd, \sv_K)]$, a quantity that is intractable to compute exactly.
As defined in \eqref{eq:bellman_a}, 
all rewards are deterministic except the rewards obtained for the last action. A straightforward way to estimate $\Ex_{\mathbf{d}\sim \mathbf{D}}[P(\bd, \sv_K)]$ is using $\rsamples$ i.i.d. samples as 
\begin{equation}\label{eq:rewardwithsamples}
\Ex_{\mathbf{d}\sim \mathbf{D}}[P(\bd, \sv_K)] \approx \frac{1}{\rsamples}\sum_{1 \leq \omega \leq \rsamples} P(\bd^{\omega}, \sv_K).   \end{equation}
\vspace{-3mm}

Intuitively, a higher value of $\rsamples$ implies less variance and gives better guidance for RL training, $i.e.$, better sample efficiency. In Section \ref{sec:ablation}, we show that the right choice of $\rsamples$ is important to the performance of the RL algorithm. 
When $\rsamples=1$, the rewards are very noisy and the RL converges slowly or even fails to converge.
If $\rsamples$ is too large, estimating $\Ex_{\mathbf{d}\sim \mathbf{D}}[P(\bd, \sv_K)]$ needs substantially more time, considering $P(\bd^\omega, \sv_K)$ is solved by 
the linear programming solver from Gurobi Optimizer \citep{gurobi}. It is interesting to see in numerical experiments that noisy rewards using a relatively small value of $\rsamples$  ($e.g.$, setting $\rsamples = 50$) can provide solutions that are similar or better compared to solutions from setting $\rsamples = 1000$ in a similar number of training steps. The observation that a small value of $\rsamples$ is enough can be explained by the effectiveness of (noisy) exploration from the literature \citep{fortunato2017noisy,plappert2017parameter}. 


\paragraph{Variance Reduction for Reward Estimation}
In addition to increasing $\rsamples$, we propose another method for reducing the variance in the reward of the MDP. In this method, instead of using the average profit of $\sv_\K$ with $\rsamples$ samples of demand as the {estimation of the} reward at the last step, we replace it by the difference of the average profits of $\flex$ and the flexibility network with all the possible arcs. Formally, 
we define $\bOne$ as a $m \times n$ matrix with all ones. Then, we change
the estimation to
\begin{equation}\label{eq:rewardwithvarreduction}
\frac{1}{\rsamples}\sum_{1 \leq \omega \leq \rsamples} \big (P(\bd^\omega, \sv_\K) - P(\bd^\omega, \bOne)\big ). 
\end{equation}
This method reduces the variance of the reward for a fixed value of $\rsamples$ because the feasible sets of the LP that defines $P(\bd, \sv_K)$ and $P(\bd, \bOne)$ both increase with $\bd$, implying that $P(\bd^\omega, \sv_K)$ and $P(\bd^\omega, \bOne)$ are positively correlated for each $\omega$. The ablation study in Section \ref{sec:ablation} shows the proposed variance reduction method helps for sample efficiency in training.

\subsection{Learning to Design via Policy Gradient}
\label{sec:training}

We considered various RL algorithms for the MDP defined in \ref{sec:MDP}, including value-based methods~\citep{mnih2013playing, wang2015dueling} and policy based method~\citep{sutton2000policy,schulman2015trust,schulman2017proximal}. Based on empirical results, we combine the advantage of policy and value based methods to learn a policy $\pi_\theta$.
Because the immediate reward released by the environment is the penalty of adding one arc except the terminal state $\sv_K$, to encourage the agent to focus on achieving better long-term rewards, we train a value network $V_{\phi}$ via
minimizing the squared residual error:
\begin{equation}
        J_{V}(\phi) = \mathbb{E}_{\tau\sim p_{\pi}(\tau)} \left( V_{\phi} (\sv_t) - \hat{V}_{\pi}(\sv_t) \right)^2\,,
\end{equation}
where $\hat{V}_{\pi}(\sv_t):= \sum_{t'=t}^K \gamma^{t'-t} r(\sv_{t'}, \av_{t'})$ is the discounted rewards-to-go. 
We further define the advantage function $A^\pi(\sv_t, \av_t)$ at step $t$ 
as \begin{align}\label{eq:advantage_fun}
A^\pi(\sv_t, \av_t) = Q(\sv_t,\av_t) - V_\phi(\sv_t)\,,
\end{align}
where $Q(\sv_t,\av_t) := r(\sv_t, \av_t) + \mathbb{E}_{{\tau}\sim p_\pi(\tau)} V_\phi(\sv_{t+1})$. 
The advantage function of a policy $\pi$ describes the gain in the expected return when choosing action $\av_t$ over randomly selecting an action according to $\pi(\cdot|\sv_t)$. We would like to maximize the expected return in \eqref{eq:rlobj} via policy gradient:
\begin{align}
{\nabla}_\theta J(\pi_\theta) = \mathbb{E}_{\tau\sim p_\pi(\tau)}\left[\sum_{t=0}^{K}A^\pi(\sv_t, \av_t)\nabla \log\pi_{\theta}(\sv_t,\av_t)\right] \,.
\end{align}
As the vanilla policy gradient is unstable, we use proximal policy optimization (PPO) with a clipped advantage function. Details of our implementation are provided in appendix~\ref{sec:ppodetail}.
%


\vspace{-1mm}
\paragraph{Policy Evaluation and FDP solutions} 
Following the literature \citep{JG95}, we design our evaluation to leverage the fact that $P(\bd, \flex)$ can be computed quickly for a given $\bd$ and $\flex$. The performance of a network is estimated using the objective function of FDP through 5000 random demand samples. Please note that using a very large number of samples in training does not provide better solution and results in a much longer training time. To reduce the training time, we adopt an early stopping strategy when there is no performance increase for 48000 steps. 

\begin{wrapfigure}{R}{0.58\textwidth}
	\begin{minipage}{0.58\textwidth}
		\vspace{-6mm}
\begin{algorithm}[H]
    \caption{Meta-Learning for FDPs}
    \label{alg1}
\begin{algorithmic}[1]
    \STATE Input: initialize parameters $\theta$ and $\phi$
    \WHILE{not done}{}
    \STATE Sample batch of tasks $\mathcal{T}_i \sim p(\mathcal{T})$
    \FOR{All $\mathcal{T}_i$}
    \STATE Sample trajectories $\tau_{\mathcal{T}_i}=\{(\sv_0, \av_0, \ldots, \sv_{K_{\mathcal{T}_i}})\}$ with $\pi_\theta$.
    \STATE Update adapted parameters $\theta_{\mathcal{T}_i}$ and $\phi_{\mathcal{T}_i}$ with $\tau_{\mathcal{T}_i}$.
    \STATE Sample trajectories $\tau=\tau \cup \{(\sv_0, \av_0, \ldots, \sv_{K_{\mathcal{T}_i}})\}$ with $\pi_{\theta_{\mathcal{T}_i}}$. 
    \ENDFOR
    \STATE Update the meta-policy $\pi_\theta$ and $V_\phi$ with $\tau$.
    \ENDWHILE
\end{algorithmic}
\end{algorithm}
	\end{minipage}
	\vspace{-2mm}
\end{wrapfigure}

\vspace{-2mm}
\paragraph{Meta-Learning with Fast Adaptation}

The ability to obtain solutions efficiently for multiple FDP instances is important in real-world applications. For example, a firm may need to solve a series of FDP instances with different $\K$, before making strategic decisions on how many arcs to have in its flexibility network. While we can apply the RL algorithm to train a different policy for each value of $\K$ from scratch, this is usually inefficient. Hence, we consider a model-agnostic meta-learning (MAML) for fast adaption~\citep{finn2017model} to take advantage of the fact that FDP instances with different $\K$ can be considered as similar tasks.

Solving FDP with different instances can be regarded as a set of tasks $\{\mathcal{T}_i\}$. The tasks $\{\mathcal{T}_i\}$ share the same deterministic transitions and similar solution spaces, and the difference between them is at most how many arcs can be added. Specifically, we first train a meta model $\pi_\theta$ using the observations from different tasks ($\K$ can be randomly selected). Then, we perform task-adaptation to adapt the meta policy to a specific task, $i.e.$, an FDP with a specific number of $K$ arcs. Intuitively, through pre-training across tasks of different $K$ values, the meta-model learns arc-adding policies that are generically effective for all the tasks, which eliminates the need for each individual task to train from scratch. The ablation study in Section \ref{sec:ablation} shows that a meta-model can quickly adapt to a specific FDP, significantly reducing the training time. In contrast, fast adaptation is not available for many classical heuristics, and the heuristics need to be performed for each task. When we have many different FDPs to solve, our meta learner presents a significant advantage in terms of efficiency and performance as shown in Section \ref{sec:ablation}.

\section{Experimental Results}\label{sec:numstudies}
We report a series of experiments conducted using the RL algorithm described in Section \ref{sec:framework}, including comparison of the RL algorithm with other heuristics under various scenarios and ablation studies.

\subsection{RL Experiment Settings}
The RL experiments are carried out on a Linux computer with one GPU of model NVIDIA GeForce RTX 2080 Ti. For all test cases, a three-layer Perceptron is used for both policy and value networks, with hidden layer sizes 1024 and 128 which is selected from ranges [256, 2048] and [64, 1024], respectively. 
For each epoch, approximately 800 episodes of games are played to collect trajectories for training. 
For the hyperparameters of the GAE, we select $\gamma = 0.99$ from the range [0.97, 0.999] and $\lambda = 0.999$ from the range of [0.97, 1]. Default values are used for the rest of hyperparameters of PPO, listed as below: 
clip ratio = 0.2, 
learning rate of $\pi$ = 3e-4,
 learning rate of the value function = 1e-3, 
train iterations for $\pi$ = 80, 
train iterations for value function =80, and 
target KL-divergence = 0.01.
Finally, for each instance of the FDP, we perform training runs with 12 different seeds, and use the trained policy network from each seed to extract 50 flexibility networks, producing a total of 600 flexibility networks. Then, we choose the best network among the 600 using a set of 5000 randomly generated demand samples.



We implement two heuristics from the literature to benchmark against the solutions from our RL algorithm. In particular, we implement the greedy heuristic and the stochastic programming based heuristic (referred to as the SP heuristic thereafter) proposed by \citet{feng2017process}. The full details for the heuristics are left in appendix \ref{sec:heuristicsdetail}. 



\subsection{RL vs Benchmark Heuristics}
We compare our RL algorithm against the two benchmark heuristics in four different scenarios. For each scenario, we create seven FDP instances by setting $\K$ equal to $\{n, n+3, \ldots, n+18\}$. We pick these values of $\K$ because if $\K<n$, then some of the demand nodes would not be connected to any arcs, creating an impractical situation where some demand nodes are being ignored completely. Also, based on the numerical study, the expected profit for all three methods reaches a plateau once $\K$ is above $n+18$. In each FDP instance, the performance for all methods, defined as the expected profit of the returned flexibility networks, are evaluated using 10,000 new randomly generated demand samples. As a benchmark, we compute an upper-bound on the optimal objective by relaxing the integrality constraints and the constraint on $K$. Next, we describe our test scenarios.

\paragraph{Automotive Manufacturing Scenario }
This scenario is motivated by an application for a large auto-manufacturer described in \citet{JG95}. In the scenario, there are 8 manufacturing plants (represented by supply nodes), where each has a fixed capacity, and 16 different vehicle models (represented by demand nodes), where each has a stochastic demand. A unique feature in this scenario is that $I_{ij}=0$ and $p_{ij}=1$ for each $i\in[m]$ and $j\in[n]$.
The goal in this scenario is to find the best production network (determining which the types of vehicle models that can be produced at each plant) with $K$ arcs that achieves the maximum expected profit.

\paragraph{ Fashion Manufacturing Scenario}  
This scenario is created based on an application for a fashion manufacturer described in \citet{chou2014performance} that extends the setting of \citet{hammond1996sport}. In this scenario, a sports fashion manufacturing firm has 10 manufacturing facilities (represented by supply nodes) and produces 10 different styles of parkas (represented by supply nodes) with stochastic demand. The goal of the firm in this scenario is to find the best production network which determines the styles of parkas that can be produced at each facility with $\K$ arcs that achieves the maximum expected profit. A unique feature in this scenario is that $I_{ij}=0$  but values of $p_{ij}$ are different for different pairs of $i$ and $j$.

\paragraph {Transportation Scenarios} 
These two test scenarios are created based on an application for a transportation firm, using publicly available testing instances on the fixed charge transportation problem (FCTP) maintained by \citet{mittelmann2005decision}. In the transportation scenarios, a firm is required to ship from origins with fixed capacities to destinations with stochastic demand, and the goal is to identify the best set of origin to destination routing network with at most $\K$ routes that achieves the maximum expected profit. The two transportation scenarios we create are based on two FCTP instances from \citet{mittelmann2005decision}, which will be referred to as transportation scenario A and transportation scenario B. In these two scenarios, values of both $I_{ij}$ and $p_{ij}$ are positive and different for different pairs of $i$ and $j$.



\begin{figure}[t!]
\centering
 \includegraphics[scale=.18]{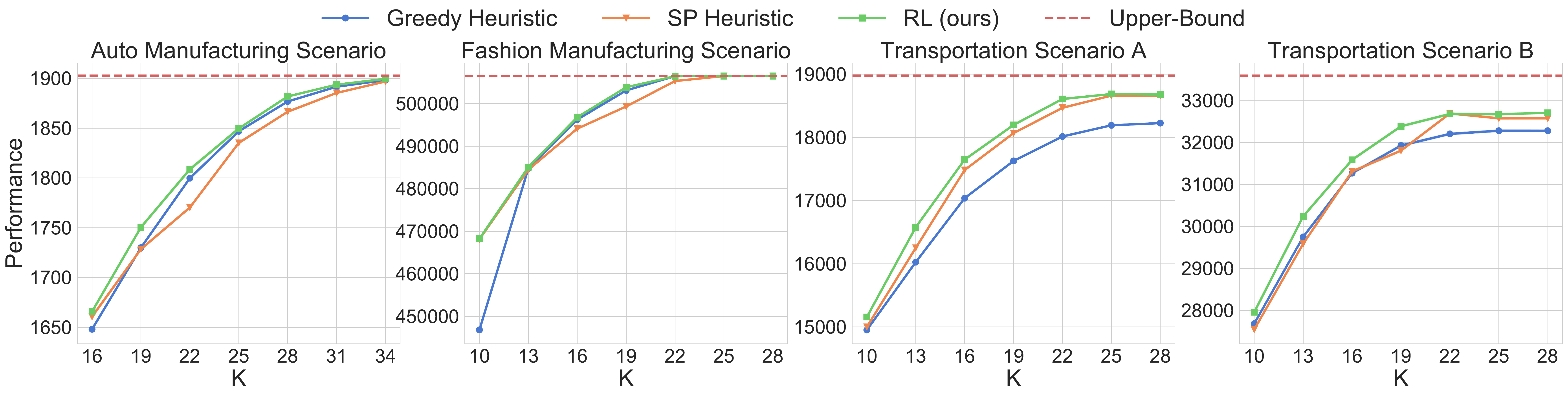}
\caption{\textit{Comparison of RL with other heuristic methods over four test scenarios.}\label{fig:compare}}
\vspace{-6mm}
\end{figure}

\paragraph{Comparison Summary Across Test Scenarios}
In Figure \ref{fig:compare}, we plot the performances of the greedy heuristic, the SP heuristic, and the RL algorithm. Figure \ref{fig:compare} demonstrates that the RL algorithm consistently outperforms both the greedy and SP heuristics.\footnote{The performance is evaluated through the empirical average over 10,000 samples, and the standard errors on the differences between the methods are less than 0.01\% for all instances.} In each of the four scenarios, the RL algorithm is better than both benchmark heuristics in the average performance over different values of $\K$. In addition, for all 28 FDP instances across the four testing scenarios, the RL algorithm is better than both benchmark heuristics in 25 of the instances, and only slightly under performs the best benchmark heuristic in the other 3 instances. Therefore, our numerical comparisons demonstrate that the RL algorithm is more effective compared to the existing heuristics. The exact performance values for the greedy heuristic, the SP heuristic, and the RL algorithm, can be found in Tables \ref{table1} to \ref{table4} in appendix \ref{sec:add_table}. In addition, we plot the training curves by plotting the average MDP return vs the number of training steps for all four testing scenarios (Figure \ref{trainingcurve} in appendix \ref{sec:add_table}). In general, the average return increases (not necessarily at a decreasing rate) with the number of training steps, until it starts to plateau. Also, the number of training steps needed to converge increases with $\K$. This is because greater $\K$ implies a longer horizon in the MDP and larger solution space, which requires a higher number of trajectories to find a good policy.

It is important to point out that our current RL algorithm has longer computational times compared to both the greedy and SP heuristics. For the test instances with $m=n=10$, the greedy heuristic takes approximately $6\times K$ seconds, the SP heuristic takes approximately $10\times K$ seconds, and the RL algorithm takes about $300\times K$ seconds. Because FDPs arise in long-term strategic planning problems, the computational time of RL is acceptable. 
For larger problems, however, the current RL algorithm may be too slow. This motivates us to perform ablation studies investigating different reward estimation and proposing meta-learning with fast adaption to reduce the training steps of RL.




\subsection{Ablation Studies}\label{sec:ablation}
We next present several ablation studies for the RL algorithm.
All ablation studies are performed under transportation scenario A, with $\K=13$ and $\K=22$. For each instance, we perform 12 training runs with different random seeds. 

\begin{figure}[htp]
\centering
  \includegraphics[scale=0.18]{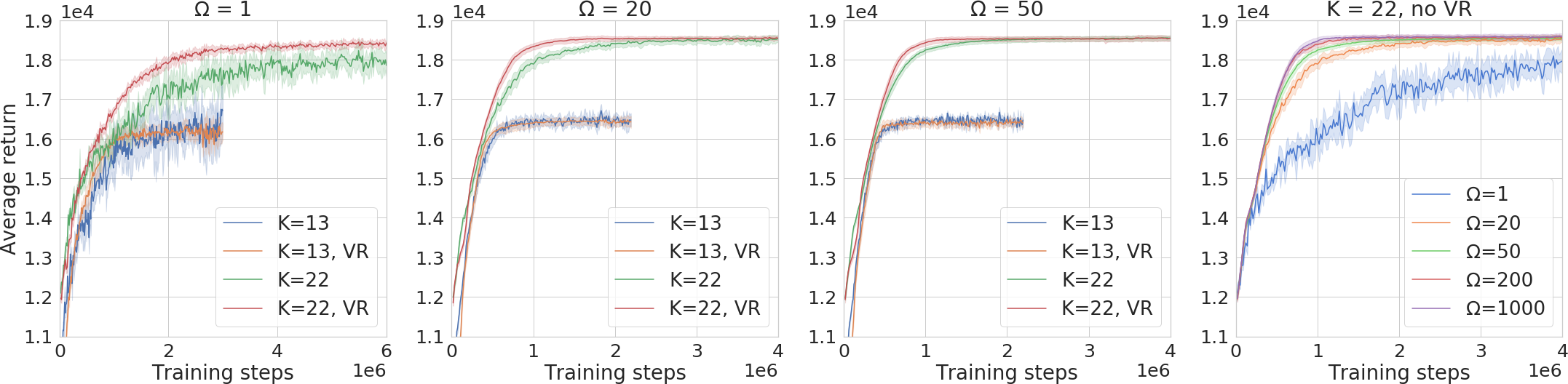}
\caption{\textit{Training curves with different values $\Omega$ with and without VR.} \label{fig:trainingcurve}}
\end{figure}

\paragraph{Using Different Reward Estimations}
In the first ablation study, we compare the RL algorithms with or without the variance reduction (VR) method described in \eqref{eq:rewardwithvarreduction}, for different values of $\rsamples$.
In Figure \ref{fig:trainingcurve}, we present the training curves for $\rsamples \in \{1,20, 50\}$ with or without the VR method, as well as training curves without the VR method for $\rsamples \in \{1,20, 50, 200, 1000\}$. We find that $\rsamples=1$ is not a good choice, as it takes many steps to converge and converges at a lower average return compared to other choices of $\rsamples$. For other values of $\rsamples$, the average return after convergence are similar, but they differ in the number of training steps required. In general, either increasing $\Omega$ or adding the VR method reduces the number of training steps. Interestingly, holding everything else equal, the reduction in training steps is smaller when $\Omega$ is large, and thus, there is little benefit in increasing $\Omega$ once in the tested instances when $\Omega$ is above 50. In addition, holding everything else equal, the reduction in train steps also increases with $\K$. This is because a larger $\K$ implies more steps in the MDP and larger solution space, and the training algorithm in such problems benefits more with smaller return variance.


\paragraph{Meta Learning with Fast Adaptation}
In the next ablation study, we test the effectiveness of meta-learning with fast adaptation. For this study, we learn a meta-model with different $\K$ with first-order MAML, either for 100 epochs (referred to as Meta100), or until convergence (referred to as MetaConvg), and then use the trained parameters from the meta-models to perform adaption for tasks with different values of $\K$.

\begin{figure}[htp]
\centering
  \includegraphics[scale=0.24]{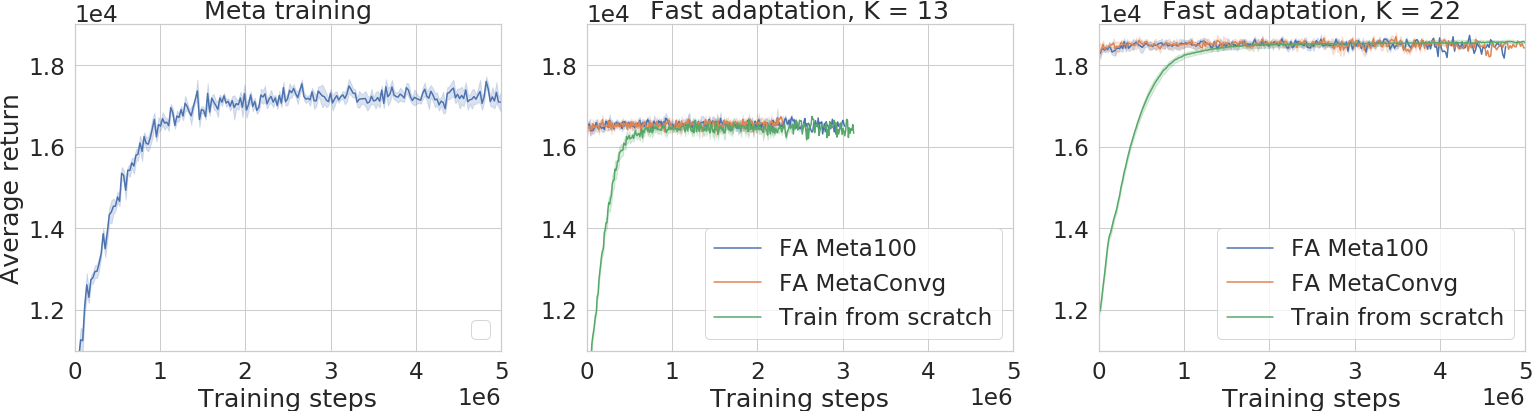}
\caption{\textit{Training curves for meta-training, and fast adaptions.}  \label{fig:metaltrainingcurve}}
\vspace{-6mm}
\end{figure}

In Figure \ref{fig:metaltrainingcurve}, we present the training curves for meta-learning and fast adaptation with both Meta100 initialization and MetaConvg initialization. We observe that meta-learning can significantly reduce the number of training steps during the fast adaption step, making it very efficient when one needs to solve FDP instances with different values of $\K$. Interestingly, we also observe that Meta100 is very much comparable to MetaConvg, implying that it is not necessary to pre-train the meta-model to convergence, thus further reducing the total number of training steps.

\section{Conclusion}

We have proposed a reinforcement learning framework for solving the flexibility design problem, a class of stochastic combinatorial optimization problems. We find that RL is a promising approach for solving FDP, and our RL algorithm consistently found better solutions compared to the benchmark heuristics. Finally, we believe that the findings of this paper suggest that an important future direction is to apply RL algorithms to other two-stage stochastic combinatorial optimization problems, especially to those problems that arise from strategic planning under uncertainty.

\bibliographystyle{iclr2021_conference}
\bibliography{rarm}
\clearpage

\appendix

\section{Details for the Benchmark Heuristics.}\label{sec:heuristicsdetail}
In the greedy heuristic, we initialize $\flex^0$ to be vector correspond to the empty flexibility network. At iteration $k$, we evaluate the expected increase in profit in adding arc $(i,j)$ to $\flex^{k-1}$ for every possible arc $(i,j)$ and 
identify arc $(i^k, j^k)$ with the highest expected profit increase. If the highest expected increase is non-positive, we terminate the algorithm and return $\flex^{k-1}$ as the final output. Otherwise, 
we add $(i^k, j^k)$ to $\flex^{k-1}$
to obtain $\flex^k$, then return $\flex^k$ as the final output if $k = \K$ or continue to iteration $k+1$. Because the expected profit for a flexibility network cannot be computed exactly, we will estimate expected profit using a set of $\Omega$ randomly generated demand samples, denoted as $\{\bd^\omega\}_{1 \leq \omega \leq \Omega}$.
More specifically, for any flexibility network $\flex$, we 
estimate its expected profit using the average profit of $\flex$ over the demand samples, that is
\begin{equation}\label{eq:sampleestimation}
\Ex_{\mathbf{d}\sim \mathbf{D}}[P(\bd, \flex)] \approx \frac{1}{\Omega}\sum_{\omega=1}^\Omega P(\bd^\omega, \flex).
\end{equation}

The SP heuristic we implement is suggested by \citet{feng2017process}. We note that although there are various heuristics proposed for more specific sub-classes of the FDP ($e.g.$, \citet{Chou2010, Chou2011, simchi2015worst}), with the exception of \citet{feng2017process}, most these heuristics are not intended for solving general FDPs. 
In the SP heuristic, we initialize $\flex^0$ to be the empty flexibility network. At iteration $k$, we consider the FDP problem that is approximated using a set of $\Omega$ randomly generated demand samples $\{\bd^\omega\}_{1 \leq \omega \leq \Omega}$, then relax its integer constraints and enforcing all arcs in $\flex^{k-1}$ to be included. In particular, we solve the
following linear stochastic programming (SP) problem:
\begin{align}
\max_{\flex\in \R^{m n},  \mathbf{f}\in \R^{mn\Omega}} \quad  & \frac{1}{\Omega}\sum_{1\leq \omega \leq \Omega}\sum_{i \in [m],j \in [n]}p_{ij}f^\omega_{ij} - \sum_{i \in [m],j \in [n]} I_{ij}F_{ij} \label{eq:spheuristic}\\
\quad \mbox{s.t.} \quad & \sum_{i \in [m]}f^\omega_{ij} \leq d^\omega_j, \, \forall j \in [n], \nonumber \\
\quad & \sum_{j \in [n]}f^\omega_{ij} \leq c_i, \, \forall i \in [m], \nonumber \\
\quad & 0 \leq f^\omega_{ij} \leq F_{ij}M_{ij} , \forall i\in[m], j\in[n], \nonumber \\
\quad & \sum_{i \in [m],j \in [n]} F_{ij} \geq F^{k-1}_{ij},\nonumber \\
\quad & \sum_{i \in [m],j \in [n]} F_{ij} \leq \K.\nonumber
\end{align}
After obtaining the solution from \eqref{eq:spheuristic}, 
the heuristic identifies arc $(i^k, j^k)$ with the highest fractional solution value among all arcs that are not in $\flex^{k-1}$, and add $(i^k, j^k)$ to $\flex^{k-1}$ to $\flex^k$. The heuristic terminates after the $\K$-th iteration. We refer to this heuristic as the \emph{SP Heuristic}.

For both greedy and SP heuristics, we choose $\Omega=1000$ and compare the networks returned by the greedy heuristic, the SP heuristic and the network returned by the RL-based algorithm for all test instances. 

\section{Details for PPO}\label{sec:ppodetail}

Our training algorithm is the proximal policy optimization (PPO) algorithm with a clipped advantage function is introduced in \citet{schulman2017proximal}. As a result, we denote our algorithm as PPO-Clip. Under the PPO-Clip algorithm, during the $k+1$-th epoch, we collect a set of trajectories $\mathcal{D}_k$ by running policy $\pi_{\theta_k}$ (a stochastic policy parameterized by $\theta_k$) in the environment $|\mathcal{D}_k|$ times, where each trajectory $\tau \in \mathcal{D}_k$ is represented by a sequence of states and actions $(s^\tau_0, a^\tau_1, s^\tau_2,, ..., s^\tau_{\K})$. After $\mathcal{D}_k$ is collected, we solve the optimization problem

\begin{equation}\label{eq:ppoclip} \max_{\theta} \sum_{\tau \in \mathcal{D}_k, t \in [\K]} \min\left(
\frac{\pi_{\theta}(a^\tau_t|s^\tau_t)}{\pi_{\theta_k}(a^\tau_t|s^\tau_t)}A^{\pi_{\theta_k}}(s^\tau_t,a^\tau_t), g(\epsilon, A^{\pi_{\theta_k}}(s^\tau_t,a^\tau_t))
\right),
\end{equation}
where $A$ is the advantage function defined in \eqref{eq:advantage_fun}, and $g$ is the clipping function defined as
\begin{align*}
g(\epsilon, A) =
\left\{
  \begin{array}{@{}ll@{}}
     (1+\epsilon)A, \, \mbox{if } A \geq 0, \\
    (1-\epsilon)A, \, \mbox{if } A < 0.
  \end{array}\right.
\end{align*}
Once \eqref{eq:ppoclip} is solved, we let $\theta_{k+1}$ to denote the optimal solution obtained in \eqref{eq:ppoclip} and update the stochastic policy by 
$\pi_{\theta_{k+1}}$. A pseudo-code for how $\pi_{\theta_{k+1}}$ is updated in our learning algorithm is provided in Algorithm \ref{alg:ppoclip}.

\begin{algorithm}[H]
    \caption{Pseudo Code for PPO-Clip}
    \label{alg:ppoclip}
\begin{algorithmic}[1]
    \STATE Input: initialize parameters $\theta_0$ and $\phi_0$
    \FOR{$k = 0,1,2,...$}
    \STATE Collect set of trajectories ${\mathcal D}$ by running policy $\pi_{\theta_k}$ in the environment.
    \STATE For each trajectory, compute the rewards-to-go $\{\hat{R}^\tau_t\}_{t=1}^T$.
    \STATE Compute advantage estimates, $\hat{A}_t$ 
    \STATE Update the policy parameter by set $\theta_{k+1}$ to be the solution of the following optimization problem:
    \begin{equation*} \theta_{k+1} = \arg\max_{\theta} \frac{1}{|{\mathcal D}_k| T} \sum_{\tau \in \mathcal{D}_k, t \in [T]} \min\left(
\frac{\pi_{\theta}(a^\tau_t|s^\tau_t)}{\pi_{\theta_k}(a^\tau_t|s^\tau_t)}\hat{A}^{\pi_{\theta_k}}(s^\tau_t,a^\tau_t), g(\epsilon, \hat{A}^{\pi_{\theta_k}}(s^\tau_t,a^\tau_t))
\right).
\end{equation*}
    \STATE Update the value function parameter by set $\phi_{k+1}$ to be the solution of the following optimization problem:
        \begin{equation*}
        \phi_{k+1} = \arg \min_{\phi} \frac{1}{|{\mathcal D}_k| T} \sum_{\tau \in {\mathcal D}_k, t\in [T]} \left( V_{\phi} (s_t) - \hat{R}^\tau_t \right)^2.
        \end{equation*}
    \ENDFOR
\end{algorithmic}
\end{algorithm}

\section{Parameters for the Test Scenarios}

\paragraph{Automotive Manufacturing Scenario}
In the automotive manufacturing scenario, the parameters for the FDP instances are set as follows.

The capacity vector for resource nodes, $\bc$, and the mean for the demand distribution at the demand nodes ($\Ex[\mathbf{D}]$), denoted by $\bmu$, are given in \citet{JG95} as 
\begin{align*}
    &\bc = [380, 230, 250, 230, 240, 230,230,240], \\
    &\bmu = [320, 150, 270, 110, 220, 110, 120, 80, 140, 160, 60, 35, 40, 35, 30, 180].
\end{align*}
We set that the distribution of $\mathbf{D}$ is independent and normal with mean $\bmu$ and standard deviation $\boldsymbol{\sigma} = 0.8\bmu$. Furthermore, for each $j$, we truncate $d_j$ to be within $[0, \mu_j + 2\sigma_j]$ to avoid negative demand instances. Finally, like \citet{JG95}, we set $p_{ij}=1$, $I_{ij}=0$ for each $1\leq i\leq m$ and $1 \leq j\leq n$.

\paragraph{Fashion Manufacturing Scenario}
In the fashion manufacturing scenario, the parameters for the FDP was selected based on the setting described in \citet{chou2014performance}. Specifically, the capacity vector for resources, $\bc$, and the mean for demand nodes, $\bmu$, are set to be equal to each other with values 
\begin{align*}
\bc = \bmu = [1017, 1042,1358,2525,1100,2150,1113,4017,3296,2383].
\end{align*}
We assume that the distribution of $\mathbf{D}$ is independent and normal with mean $\bmu$ and standard deviation $\boldsymbol{\sigma}$, where the values of $\boldsymbol{\sigma}$ is obtained from \citet{chou2014performance}, with $\boldsymbol{\sigma}= [194,323,248,340,381, 404, 524, 556, 1047, 697]$. Furthermore, for each $j$, we truncate $d_j$ to be within $[0, \mu_j + 2\sigma_j]$ to avoid negative demand instances. In addition, the prices of the parkas styles are given by
$$\boldsymbol{q} = [110, 99, 80, 90, 123, 173, 133, 73, 93, 148],$$
and the profit margin is 24\%. Thus, we set the unit profit for using one resource $i$ to satisfy one demand $j$, $p_{ij}$, to $p_{ij} = 0.24q_j$, for each $1\leq i\leq m$ and $1 \leq j\leq n$. Finally, like \citet{chou2014performance}, we choose  $I_{ij}=0$ for all $1\leq i\leq m$ and $1 \leq j\leq n$. 

\paragraph{Transportation Scenarios}
We create transportation scenario A and transportation scenario B from two FCTP instances from \citet{mittelmann2005decision}, which are named as `ran10x10a' and '`ran10x10b' in \citet{mittelmann2005decision}. In an FCTP, the firm is required to ship from origins with fixed capacities to destinations with fixed demands, and the goal is to minimize the total transportation cost plus the fixed cost of transporting on a origin-destination transportation network. To convert an FCTP into an FDP, we use the supply nodes to represent the origins and the demand nodes to represent the destinations, and perturb the demand of the destinations with a random noise. More specifically, we set the capacity vector $\bc$ as the capacity of the origins and the mean for demand nodes as the demand of the destinations from the FCTP problem. We further set that the distribution of $\mathbf{D}$ is independent and normal with mean $\bmu$ and standard deviation $\boldsymbol{\sigma} = 0.8\bmu$. For each $1\leq i\leq m$ and $1 \leq j\leq n$, the value of $I_{ij}$ is set to the fixed charge of each arc of the FCTP instance. In addition, let the unit transportation cost from the FCTP instance for each original $i$ and destination $j$ to be $t_{ij}$. We set the unit profit from resource $i$ to demand $j$, $p_{ij}$, to be a large constant $P_{const}$ minus $t_{ij}$ where $P_{const}$ is computed as 
$$P_{const} = \max_{(i,j)\in \R^{mn}} (I_{ij} + t_{ij}).$$
This choice of $P_{const}$ and $p_{ij}$ allows the FTP problem with an unbounded number of arcs to be equivalent to the original FCTP testing instance when the standard deviation of $\mathbf{D}$ is the zero vector. 



\section{Additional Ablation Study}

\paragraph{Enlarging the Action Set} 
We consider a different action set where given state $s=\flex$, each action can remove arc $(i,j)$ from $\flex$ if $F_{ij}=1$, add $(i,j)$ if $F_{ij}=0$, or choose to do nothing and keep the current state ($i.e.$, no-op). We can use this action set to replace the original action set defined in the MDP and obtain a new MDP that is also equivalent to the FDP problem. 

This action set is considered because the new MDP has more opportunities to change the state than the original action set, and thus help us to check if the additional opportunities in the new action set may fasten the training process or help our RL algorithm to find policies that identify better FDP solutions compared to the original action set. We denote the default action set as \textbf{add/no-op} and the new action set as \textbf{add/delete/no-op}.

\begin{table}[htp]
\small
\begin{center}
\caption{Average performance, best performance, and average training steps for different action sets}
\label{table:ablationaction}
\begin{tabular}{l|lll|lll}
\toprule[1.2pt]
&   & $K=13$ &  &  &$K=22$ & \\ \hline
 & Avg Perf & Best Perf & Avg Steps & Avg Perf & Best Perf & Avg Steps \\
add/no-op & 16463 & 16597 & 132640 & 18529 & 18605 & 181360 \\
add/delete/no-op & 16469 & 16670 & 154640 & 18528 & 18622 & 218640  \\
\bottomrule[1.2pt]
\end{tabular}
\end{center}
\end{table}


In Table \ref{table:ablationaction}, we present the average performance and best performance of the structures returned by the MDP with the original action set (add/no-op) and the add/delete/no-op action set under transportation scenario A with $\K=13$ and $\K=22$. We find no significant difference in the performances between the two action sets. In addition, we also provide the average number of training steps used in training for both action sets. The RL algorithm takes on average more training steps under the new action set, and therefore require a longer computational time as well. Therefore, we conclude that there seems to be no benefit to use the  add/delete/no-op action set.

\section{Additional Tables and Graphs for Experimental Results}\label{sec:add_table}
\begin{figure}[htp]
\centering  
\includegraphics[scale=0.18]{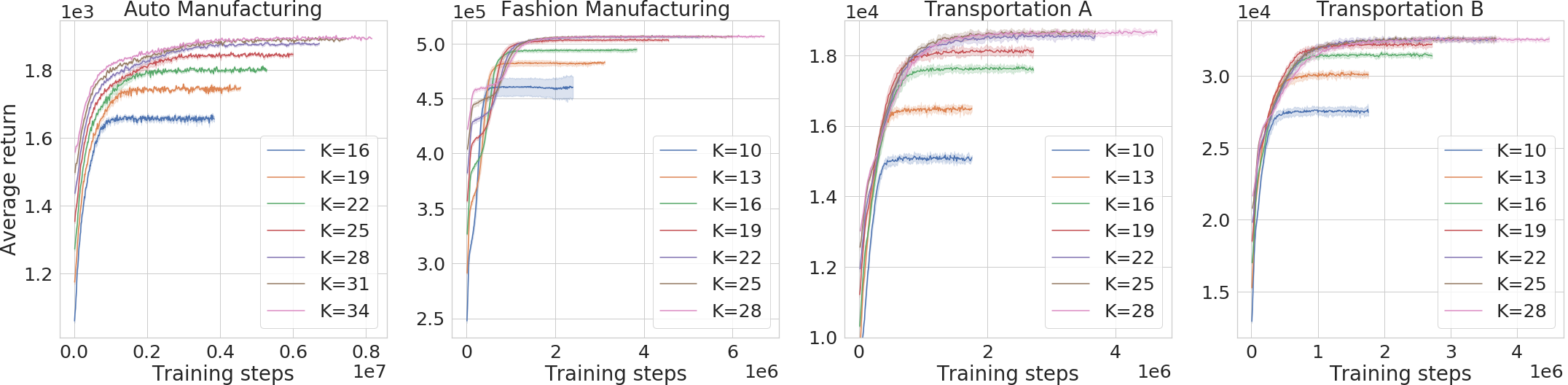}
\caption{\textit{Training curve for different values of $\K$.}  \label{trainingcurve}}
\vspace{-6mm}
\end{figure}

We provide the specific values for the expected profit achieved by each of the methods for all of the test scenarios, via Tables \ref{table1} thru \ref{table4}. For each instance, the method with the highest expected profit is highlighted in bold. The training curves for RL for all test scenarios are presented as Figure \ref{trainingcurve}

\begin{table}[htp]
\begin{center}
\caption{Expected Profit for RL vs Greedy and SP Heuristics for the Auto Manufacturing Scenario}
\begin{tabular}{lrrrrrrr}
\toprule[1.2pt]
\textbf{Methods}                   & $K$\textbf{=16}    & \textbf{19}      & \textbf{22}      & \textbf{25}      & \textbf{28}      & \textbf{31}      & \textbf{34}       \\ \hline
Greedy & 1648.0 & 1730.0 & 1799.8 & 1846.9 & 1876.8 & 1891.6 & 1898.3 \\
SP & 1660.7 & 1728.5 & 1770.3 & 1835.2 & 1866.5 & 1885.4 & 1896.9 \\
RL & \textbf{1665.7} & \textbf{1750.5} & \textbf{1808.7} & \textbf{1849.7} & \textbf{1881.8} & \textbf{1893.7} & \textbf{1899.6} \\\hline
\bottomrule[1.2pt]
\end{tabular}
\label{table1}
\end{center}
\end{table}

\begin{table}[htp]
\begin{center}
\caption{Expected Profit for RL vs Greedy and SP Heuristics for the Fashion Manufacturing Scenario \label{table2}}
\begin{tabular}{lrrrrrrr}
\toprule[1.2pt]
\textbf{Methods} 
& $\K$ \textbf{= 10} & \textbf{13} & \textbf{16} & \textbf{19} & \textbf{22} & \textbf{25} & \textbf{28} \\ \hline
Greedy & 446809.9 & 484788.8 & 496262.8 & 503107.5 & \textbf{506480.3} & 506497.2 & 506497.2 \\
SP & 468137.4 & 484474.0 & 494125.4 & 499337.6 & 505321.3 & \textbf{506499.9} & 506500.1 \\
RL & \textbf{468248.3} & \textbf{485085.0} & \textbf{496812.6} & \textbf{503852.2} & 506449.5 & 506497.0 & \textbf{506501.3} \\ \hline
\\
\bottomrule[1.2pt]
\end{tabular}
\end{center}
\end{table}




\begin{table}[htp]
\begin{center}
\caption{Expected Profit for RL vs Greedy and SP Heuristics for Transportation Scenario 3A\label{table3}}
\begin{tabular}{lrrrrrrr}
\toprule[1.2pt]
\textbf{Methods} 
& $\K$ \textbf{= 10} & \textbf{13} & \textbf{16} & \textbf{19} & \textbf{22} & \textbf{25} & \textbf{28} \\ \hline
Greedy & 14950.8 & 16025.0 & 17038.3 & 17627.4 & 18014.4 & 18191.6 & 18226.3 \\
SP & 14999.2 & 16250.4 & 17483.7 & 18067.6 & 18468.4 & 18661.4 & 18661.8 \\
RL & \textbf{15155.6} & \textbf{16577.9} & \textbf{17648.3} & \textbf{18197.1} & \textbf{18608.0} & \textbf{18686.4} & \textbf{18678.7}\\ \hline
\bottomrule[1.2pt]
\end{tabular}
\end{center}
\end{table}

\begin{table}[h!]
\begin{center}
\caption{Expected Profit for RL vs Greedy and SP Heuristics for Transportation Scenario 3B\label{table4}}
\begin{tabular}{lrrrrrrr}
\toprule[1.2pt]
\textbf{Methods} 
& $\K$ \textbf{= 10} & \textbf{13} & \textbf{16} & \textbf{19} & \textbf{22} & \textbf{25} & \textbf{28} \\ \hline
Greedy & 27682.5 & 29751.5 & 31271.1 & 31929.1 & 32207.2 & 32282.6 & 32282.6 \\
SP & 27546.1 & 29587.2 & 31312.2 & 31808.0 & \textbf{32695.0} & 32578.0 & 32578.0 \\
RL & \textbf{27956.2} & \textbf{30241.9} & \textbf{31588.1} & \textbf{32388.5} & 32682.5 & \textbf{32674.8} & \textbf{32710.1} \\ \hline
\bottomrule[1.2pt]
\end{tabular}
\end{center}
\end{table}

\end{document}

%% file: math_commands.tex
\usepackage{amsmath,amsfonts,bm}

\def\eqref#1{equation~\ref{#1}}

\def\1{\bm{1}}

\DeclareMathAlphabet{\mathsfit}{\encodingdefault}{\sfdefault}{m}{sl}
\SetMathAlphabet{\mathsfit}{bold}{\encodingdefault}{\sfdefault}{bx}{n}

\newcommand{\R}{\mathbb{R}}

\usepackage{hyperref}
\usepackage{url}

\newcommand{\bc}{\boldsymbol{c}}

\newcommand{\bd}{\boldsymbol{d}}

\newcommand{\bp}{\boldsymbol{p}}

\newcommand{\bOne}{\mathbf{1}}

\newcommand{\bmu}{\boldsymbol{\mu}}

\newcommand{\K}{K} 
\newcommand{\Ex}{\mathbb{E}}
\newcommand{\flex}{\mathbf{F}}
\newcommand{\rsamples}{\Omega}

\usepackage{algorithm}
\usepackage{algorithmic}

\usepackage{amsmath}
\usepackage{amsfonts}
\usepackage{amssymb}

\newcommand{\beq}{\vspace{0mm}\begin{equation}}
\newcommand{\eeq}{\vspace{0mm}\end{equation}}
\newcommand{\beqs}{\vspace{0mm}\begin{eqnarray}}
\newcommand{\eeqs}{\vspace{0mm}\end{eqnarray}}
\newcommand{\barr}{\begin{array}}
\newcommand{\earr}{\end{array}}
\newcommand{\av}{{\boldsymbol a}}

\newcommand{\sv}{{\boldsymbol s}}

\usepackage[bottom]{footmisc}

\def\bp{\mathbf{p}}